\documentclass[11pt]{article}

\usepackage{ifxetex,ifluatex}
\ifnum 0\ifxetex 1\fi\ifluatex 1\fi=0
  \usepackage[utf8]{inputenc}
  \usepackage[T1]{fontenc}
\fi
\usepackage{amsmath,amssymb,amsfonts}
\usepackage{graphicx}
\usepackage{algorithm}
\usepackage{algorithmic}
\usepackage{booktabs}
\usepackage{url}
\usepackage{hyperref}
\usepackage{fontspec}
\usepackage{tabularx}

\usepackage{tikz}
\usetikzlibrary{shapes.geometric, arrows.meta, positioning, calc, fit, backgrounds, decorations.pathreplacing}
\usepackage{pgfplots}
\pgfplotsset{compat=1.18}

\definecolor{umfblue}{RGB}{41, 128, 185}
\definecolor{umfgreen}{RGB}{39, 174, 96}
\definecolor{umforange}{RGB}{230, 126, 34}
\definecolor{umfred}{RGB}{192, 57, 43}
\definecolor{umfpurple}{RGB}{142, 68, 173}
\definecolor{umfgray}{RGB}{127, 140, 141}

\newcommand{\sintext}[1]{{\fontspec{iskola-pota.ttf}#1}}

\usepackage[letterpaper, margin=1in]{geometry}

\title{Typologically-Informed Candidate Reranking for LLM-based Translation into Low-Resource Languages}

\author{ \parbox{\textwidth}{\centering\textbf{Nipuna Abeykoon$^{*}$, Ashen Weerathunga$^{*}$, Pubudu Wijesinghe$^{*}$, Parameswari Krishnamurthy}} \\ \\ 
ZWAG AI Ltd, Dubai AI Campus, DIFC, UAE. \\ \\ 
\parbox{\textwidth}{\centering\small $^{*}$These authors contributed equally to this work} 
}

\date{}

\begin{document}

\maketitle

\begin{abstract}
Large language models trained predominantly on high-resource languages exhibit systematic biases toward dominant typological patterns, leading to structural non-conformance when translating into typologically divergent low-resource languages. We present a framework that leverages linguistic typology to improve translation quality without parallel training data or model retraining. The framework consists of two components: the Universal Metalinguistic Framework (UMF), which represents languages as structured profiles across 16 typological dimensions with divergence-weighted scoring, and the Computational Engine, which operates through linguistic disambiguation during generation and typological compliance scoring during selection.
Evaluation across nine language pairs demonstrates intervention rates strongly correlating with typological distance from English. In experiments on 341 English sentences each having different morphological and syntactic phenomena, the framework shows an intervention precision of 48.16\% for conservatively treated languages, 28.15\% for morphologically dense languages, and 86.26\% for structurally profiled languages. The framework requires no parallel training data and operates with any LLM capable of producing multiple candidate outputs, enabling practical deployment for under-resourced languages.

\end{abstract}


\section{Introduction}

Modern LLMs treat language as surface-level data, not as a cognitive system.  As a result, their performance is optimized for languages whose grammatical structures, cultural assumptions, and reasoning patterns dominate training corpora. Languages with different structural properties are frequently distorted, flattened, or misrepresented.

Large language models learn statistical regularities from datasets that are overwhelmingly English-centric. When applied to structurally distinct languages, these models often preserve semantic intent but fail to maintain structural correctness. This leads to systematic errors in tense, agency, evidentiality, politeness, spatial relations, and relational meaning. Outputs may appear fluent yet remain cognitively incongruent for native speakers.

We term these failures \textit{interpretation errors}: outputs that preserve surface semantic content but violate obligatory grammatical, morphosyntactic, or discourse constraints required for correct interpretation by native speakers. Unlike stylistic variations or acceptable paraphrases, interpretation errors produce translations that are grammatically malformed, pragmatically inappropriate, or cognitively dissonant in the target language, even when the intended meaning remains recoverable.

This is not merely a matter of data scarcity. Human languages exhibit extraordinary structural diversity: word order patterns range from Subject-Verb-Object (English) to Subject-Object-Verb (Hindi, Japanese) to Verb-Subject-Object (Arabic); morphological systems span from analytic (Chinese) to agglutinative (Turkish, Swahili) to polysynthetic (Inuktitut). Case marking, agreement patterns, honorifics, and dozens of other grammatical dimensions vary independently across languages~\cite{ref9}. Models trained predominantly on English must somehow produce structurally valid outputs for languages exhibiting fundamentally different typological patterns. This is a challenge that data volume alone cannot solve. The result is systematic structural errors: incorrect word order, missing morphological markers, inappropriate lexical choices, and violations of target language grammatical constraints. This inequality stems from fundamental biases in how these models are trained and evaluated.

The root cause is typological bias: models trained predominantly on English encode structural patterns of Subject-Verb-Object word order, analytic morphology, and minimal case marking. When translating to typologically divergent languages, these biases manifest as systematic errors across multiple dimensions.

Structural errors arise from grammatical mismatches. An English-to-Sinhala translation of the sentence ``The children play in the garden'' may preserve SVO order instead of restructuring to Sinhala's SOV pattern, producing \sintext{දරුවන් සෙල්ලම් උද්‍යානය} rather than the correct \sintext{ළමයි උද්‍යානයේ සෙල්ලම් කරනවා}. The model may omit obligatory case markers: Sinhala requires the locative suffix \sintext{-ේ} (\textit{-ē}) on ``garden'' to indicate location or fail to produce appropriate morphological complexity. Recent analyses confirm this pattern: language models achieve significantly higher scores on fusional languages but underperform on agglutinative languages due to richer morphology and greater token sparsity~\cite{ref3}.

Lexical errors stem from training distribution biases. In our experiments with frontier LLMs, translating "play" yielded \sintext{වාදනය} (vādanaya, "play a musical instrument") rather than \sintext{සෙල්ලම්} (sellam, "play/have fun") because the former appears more frequently in English-Sinhala training data despite being contextually inappropriate for "children playing in a garden."

Existing remedies have significant limitations. Fine-tuning on parallel corpora requires data that does not exist for most language pairs~\cite{ref8}, while prompt engineering lacks precision to enforce complex grammatical constraints~\cite{ref6, ref7}. Prior work has incorporated typological insights in limited or auxiliary forms; however, no existing system operationalizes typology as a universal, structured decision layer that directly governs inference-time candidate selection across languages without retraining or parallel data. This motivates our approach: leverage explicit typological knowledge to guide translation under black-box generation.

The Universal Metalinguistic Framework (UMF) addresses this challenge by quantifying typological divergence between languages and using it to guide translation without retraining. The framework targets both error types through complementary mechanisms. For lexical errors, a semantic constraint layer applies context-aware adjustments during generation to resolve word-sense ambiguities. For structural errors, a typological reranking layer evaluates candidates against explicit grammatical requirements of the target language.

UMF represents languages through expert-curated profiles capturing 16 typological dimensions based on the World Atlas of Language Structures~\cite{ref9} and internal linguistic typology research. Divergence scores quantify how much source and target languages differ in each dimension. These scores are weighted by linguistic importance and normalized to produce a directive vector that guides candidate evaluation. A tunable mixing parameter $\alpha$ balances the model's probability with the typological compliance score.

Our approach is grounded in the principle that typological divergence predicts where LLMs will fail. Languages with divergent word order, rich case systems, and agglutinative morphology receive higher intervention rates, with the reranker prioritizing candidates that satisfy target language constraints. This dual-layer approach, combining typological scoring with semantic disambiguation, yields improvements that correlate with typological divergence and remain consistent across different base models.

\textbf{Scope and Non-Claims.} To clarify the boundaries of this work: UMF does not generate translations; it selects among candidates produced by an existing model. UMF does not replace automatic evaluation metrics; it operates as a complementary decision layer for candidate selection. UMF does not claim universal coverage—current language profiles cover a subset of typological phenomena and require expansion. What UMF does provide is a structured, interpretable mechanism for enforcing typological constraints at inference time, operating as a metalinguistic decision layer rather than a learned quality predictor.

\section{Background and Related Work}

\subsection{Typology in Metalinguistic and MT Frameworks: Prior Approaches}
Linguistic typology has been incorporated into machine translation and metalinguistic frameworks in various forms, yet these approaches have treated typological knowledge as auxiliary, analytical, or partial rather than as a foundational decision-governing component. Theoretical models such as Universal Grammar (UG) were built on the assumption of deep structural uniformity across languages, reducing variation to a small set of abstract principles or binary parameters. Large-scale typological evidence has since demonstrated that linguistic diversity is far richer and less discretizable than UG predicts, with many languages exhibiting structural configurations that fall outside proposed universals \cite{ref15, ref16}. Interlingua-based systems such as ATLAS pursued language-independent representations, yet empirical evaluations showed that no interlingua could remain neutral across languages without reintroducing language-specific transfer rules, effectively collapsing back into pairwise or family-specific solutions \cite{ref17, ref18}. Rule-based MT systems encoded typological structure explicitly but required exhaustive manual specification of grammatical rules, making them incomplete, non-scalable, and biased toward well-documented Indo-European language structures \cite{ref19}.

Data-driven paradigms, including statistical and neural machine translation, improved surface fluency and scalability but introduced a different class of limitations. By learning exclusively from observed corpora, these models implicitly privilege high-resource languages and frequent constructions, systematically underrepresenting rare, optional, or typologically marked phenomena such as rich morphology, evidentiality, honorific systems, or non-configurational syntax \cite{ref20, ref21}. Neural MT models, in particular, optimize for probabilistic fluency rather than structural faithfulness, often omitting or flattening linguistic features that lack clear cross-lingual alignment or sufficient training signal \cite{ref22}. While typological information has been incorporated in prior frameworks, including divergence indices for transfer-based MT \cite{ref14} and typological feature embeddings, it has remained auxiliary, partial, or analytical rather than serving as a structured, decision-governing component of translation systems \cite{ref23}. As a result, modern MT systems remain effective pattern translators rather than linguistically grounded interpreters, with typological knowledge functioning as metadata or side-channel features rather than as a first-class evaluation mechanism that directly governs inference-time selection.

\subsection{Typological Bias in LLMs}

Linguistic typology classifies languages along structural dimensions that vary independently. Morphological typology distinguishes analytic, agglutinative, fusional, and polysynthetic systems. Analytic languages (e.g., Mandarin) use minimal inflection and rely on word order and particles. Agglutinative languages (e.g., Turkish, Japanese, Swahili, Tamil) construct words by concatenating discrete morphemes, each encoding a single grammatical meaning. A noun may carry suffixes for case, number, and possession in sequence. Fusional languages (e.g., Spanish, Russian) compress multiple grammatical categories into single portmanteau morphemes. Word order typology captures the positioning of subject (S), verb (V), and object (O), with SVO, SOV, and VSO accounting for the vast majority of languages~\cite{ref9}.
Large language models disproportionately favor analytic and fusional languages. Arnett and Bergen~\cite{ref3} found a performance gap between agglutinative and fusional languages, with fusional languages such as English achieving lower perplexities, attributed partly to tokenization and dataset size disparities. Brinkmann et al.~\cite{ref2} observe that leading multilingual models remain primarily English models, with over 90\% of training tokens in English. The authors hypothesize that while LLMs learn some language-invariant abstractions, these abstractions may still be biased toward English grammar and semantics. Such observations motivate our efforts to explicitly encode typological properties into the translation process.

\subsection{Low-Resource Machine Translation}

Transfer learning exploits related high-resource languages to bootstrap low-resource translation. Boujkian et al.~\cite{ref8} show that linguistic similarity enables efficient adaptation, but this method requires parallel data and yields diminishing returns as typological distance increases. Fine-tuning on in-domain parallel corpora remains standard but is unavailable for thousands of languages lacking digital corpora.
Krishnamurthy~\cite{ref14} addresses typological divergence in Telugu-Tamil transfer-based MT through a Divergence Index (DI) that quantifies linguistic differences across five levels, namely, surface, shallow, intermediate, deep, and deeper. The DI enabled targeted improvements in transfer grammar rules, increasing fluency from 63\% to 87\%. While our approach differs architecturally (reranking vs. transfer rules), both frameworks ground translation improvement in explicit typological divergence measurement. Krishnamurthy's work demonstrates that quantifying linguistic distance enables systematic error correction, which is a principle central to our UMF scoring mechanism.

\subsection{Candidate Reranking and Self-Evaluation}

Reranking methods select among multiple candidates rather than relying on the model's top output. Traditional N-best list reranking incorporated linguistic features to re-score candidates. Recent LLM-as-a-judge approaches use models to evaluate their own outputs. Franceschelli and Musolesi~\cite{ref10} propose Creative Beam Search, using diverse beam search to generate varied candidates and self-evaluation to select outputs, addressing positional bias through balanced position calibration. While effective for subjective criteria, this approach lacks grounding in explicit linguistic constraints and cannot systematically correct structural errors like word order violations or missing morphological markers.

\subsection{Quality Estimation}

Automatic metrics like BLEU~\cite{ref11} measure n-gram overlap but are insensitive to structural correctness. Neural metrics like BERTScore and COMET~\cite{ref12, ref13} achieve stronger correlation with human judgments but may not reliably detect typological errors for underrepresented languages. Critically, all metrics predict quality but do not identify or correct errors.

\subsection{Controllable Generation}

Prompt engineering provides high-level instructions but lacks precision to enforce complex grammatical constraints. Constrained decoding methods enforce hard lexical constraints but cannot express structural requirements like ``apply SOV ordering.'' Logit-level interventions (PPLM, FUDGE) modify token probabilities during generation but require model internals access and cannot reason about sentence-level structural properties.

Our approach differs in three key aspects: we operate on complete candidates (black-box compatible), ground evaluation in explicit typological profiles (not learned correlations), and target both lexical and structural errors. Crucially, UMF does not learn correctness from data—it encodes grammatical obligation from linguistic knowledge. This distinction is fundamental: learned quality estimators predict what outputs humans prefer, while UMF enforces what outputs the target language requires.

\section{The UMF Framework and Computational Engine}

\subsection{Overview}

The Universal Metalinguistic Framework consists of two main components:

\textbf{Universal Metalinguistic Framework (UMF):} A structured representation of language typology encoding 16 dimensions derived from linguistic research. Each language is described by a profile specifying its typological characteristics. Divergence scores quantify differences between source and target languages along each dimension.

\textbf{Computational Engine:} Implements candidate generation, scoring, and reranking. The engine operates in two phases: (1) Semantic Constraint Layer applies context-aware lexical adjustments during generation, and (2) Typological Scoring evaluates candidates for structural compliance with the target language.

The framework is model-agnostic and operates on any LLM capable of producing multiple translation candidates. It requires no fine-tuning or parallel training data, only a typological profile for the target language.

\begin{figure}[htbp]
\centering
\begin{tikzpicture}[
    node distance=0.8cm,
    box/.style={rectangle, draw, rounded corners, minimum width=2.8cm, minimum height=0.9cm, align=center, font=\small},
    mainbox/.style={box, fill=umfblue!20, draw=umfblue, line width=1pt},
    layerbox/.style={box, fill=umfgreen!20, draw=umfgreen},
    scorebox/.style={box, fill=umforange!20, draw=umforange},
    outputbox/.style={box, fill=umfpurple!20, draw=umfpurple},
    profilebox/.style={box, fill=umfgray!20, draw=umfgray},
    arrow/.style={-{Stealth[length=2.5mm]}, thick},
    groupbox/.style={rectangle, rounded corners=5pt, draw, dashed, line width=1.2pt, inner sep=10pt}
]

\node[mainbox] (input) {Source Text\\``The children play''};

\node[mainbox, below=of input] (llm) {LLM Generator\\(GPT-5.2 / mT5)};

\node[mainbox, below=of llm] (candidates) {N Candidates\\(Beam Search)};

\node[profilebox, right=2.5cm of llm] (srcprof) {Source Profile\\(English)};
\node[profilebox, below=0.4cm of srcprof] (tgtprof) {Target Profile\\(Sinhala)};

\node[scorebox, below=0.4cm of tgtprof] (directive) {Directive Engine\\16D Vector};

\begin{scope}[on background layer]
    \node[groupbox, draw=umforange, fill=umforange!5, fit=(srcprof)(tgtprof)(directive), label={[font=\small\itshape, text=umforange]above:Universal Metalinguistic Framework}] (umfgroup) {};
\end{scope}

\node[layerbox, below=1.2cm of candidates, xshift=-1.8cm] (semantic) {Layer 1:\\Semantic\\Constraints};
\node[layerbox, below=1.2cm of candidates, xshift=1.8cm] (typological) {Layer 2:\\Typological\\Scoring};

\node[scorebox, below=0.9cm of semantic, xshift=1.8cm] (scoring) {Combined Score\\$\alpha \cdot P_{model} + (1-\alpha) \cdot UMF$};

\begin{scope}[on background layer]
    \node[groupbox, draw=umfgreen, fill=umfgreen!5, fit=(semantic)(typological)(scoring), label={[font=\small\itshape, text=umfgreen]below:Computational Engine}] (cegroup) {};
\end{scope}

\node[outputbox, below=1.2cm of scoring] (output) {Best Translation\\\sintext{ළමයි සෙල්ලම් කරනවා}};

\draw[arrow] (input) -- (llm);
\draw[arrow] (llm) -- (candidates);
\draw[arrow] (candidates) -- (semantic);
\draw[arrow] (candidates) -- (typological);
\draw[arrow] (semantic) -- (scoring);
\draw[arrow] (typological) -- (scoring);
\draw[arrow] (scoring) -- (output);

\draw[arrow, dashed, umfgray] (srcprof) -- (tgtprof);
\draw[arrow, dashed, umfgray] (tgtprof) -- (directive);
\draw[arrow, dashed, umforange] (directive) -| (typological);

\end{tikzpicture}
\caption{UMF Framework Architecture. The source text is processed through an LLM to generate N candidates. The Universal Metalinguistic Framework (right, orange-dashed) computes a 16-dimensional directive vector from source and target language profiles. The Computational Engine (center, green-dashed) applies dual-layer evaluation: semantic constraints for lexical disambiguation and typological scoring for structural compliance.}
\label{fig:architecture}
\end{figure}
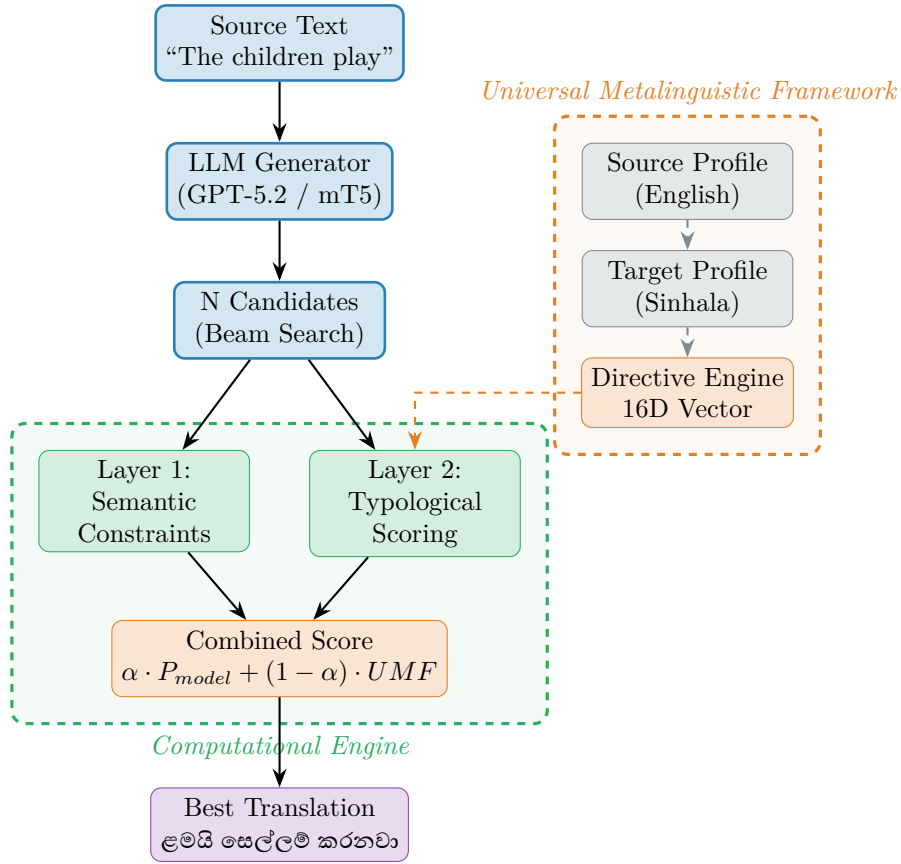

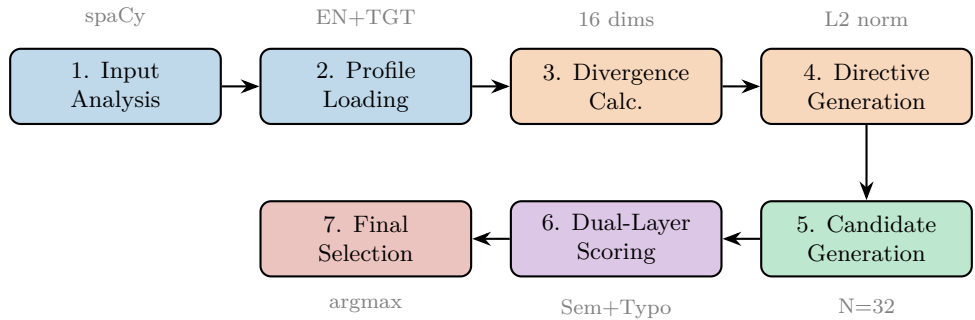
\begin{figure}[htbp]
\centering
\begin{tikzpicture}[
    node distance=0.5cm and 0.5cm,
    stagebox/.style={rectangle, draw, rounded corners=4pt, minimum width=2.8cm, minimum height=1.0cm, align=center, font=\footnotesize, line width=0.8pt},
    myarrow/.style={-{Stealth[length=2.5mm]}, thick},
    smalllabel/.style={font=\scriptsize, text=gray}
]

\node[stagebox, fill=umfblue!30] (s1) {1. Input\\Analysis};
\node[stagebox, fill=umfblue!30, right=of s1] (s2) {2. Profile\\Loading};
\node[stagebox, fill=umforange!30, right=of s2] (s3) {3. Divergence\\Calc.};
\node[stagebox, fill=umforange!30, right=of s3] (s4) {4. Directive\\Generation};

\node[stagebox, fill=umfgreen!30, below=1.0cm of s4] (s5) {5. Candidate\\Generation};
\node[stagebox, fill=umfpurple!30, left=of s5] (s6) {6. Dual-Layer\\Scoring};
\node[stagebox, fill=umfred!30, left=of s6] (s7) {7. Final\\Selection};

\draw[myarrow] (s1) -- (s2);
\draw[myarrow] (s2) -- (s3);
\draw[myarrow] (s3) -- (s4);

\draw[myarrow] (s4) -- (s5);

\draw[myarrow] (s5) -- (s6);
\draw[myarrow] (s6) -- (s7);

\node[smalllabel, above=0.15cm of s1] {spaCy};
\node[smalllabel, above=0.15cm of s2] {EN+TGT};
\node[smalllabel, above=0.15cm of s3] {16 dims};
\node[smalllabel, above=0.15cm of s4] {L2 norm};
\node[smalllabel, below=0.15cm of s5] {N=32};
\node[smalllabel, below=0.15cm of s6] {Sem+Typo};
\node[smalllabel, below=0.15cm of s7] {argmax};

\end{tikzpicture}
\caption{UMF Translation Pipeline. The seven-stage process transforms source text into typologically-compliant translations through linguistic analysis, divergence quantification, multi-candidate generation, and dual-layer evaluation.}
\label{fig:pipeline}
\end{figure}

\subsection{Language Profile Structure}

A language profile is a structured representation capturing typological properties. Each profile consists of 16 typological dimensions representing structural characteristics. Each dimension includes:
\begin{itemize}
\item \textbf{Value:} The language's specific property for that dimension (e.g., word order = SOV).
\item \textbf{Weight:} Linguistic importance of the dimension, derived from typological research indicating how strongly the dimension influences grammatical structure.
\item \textbf{Markers:} Observable surface features used for evaluation (e.g., case suffixes, verb endings).
\end{itemize}

Language profiles are expert-curated JSON structures encoding 16 typological dimensions derived from the World Atlas of Language Structures \cite{ref9} and linguistic typology research. Each dimension captures a structural property with cross-linguistic variation. Table \ref{tab:dimensions} lists all dimensions with brief descriptions.

\begin{table}[htbp]
\centering
\caption{The 16 Typological Dimensions}
\label{tab:dimensions}
\small
\begin{tabularx}{\textwidth}{lXX}
\toprule
\textbf{Dimension} & \textbf{Description} & \textbf{Example Contrast} \\
\midrule
Word order & Canonical constituent ordering & SVO (English) vs. SOV (Sinhala) \\
Case marking & Grammatical case inventory and marking & 3 cases (English) vs. 8 cases (Sinhala) \\
Morphology & Morphological type and complexity & Analytic (English) vs. Agglutinative (Sinhala) \\
Agreement & Subject-verb and noun-adjective patterns & Minimal (English) vs. Rich (Sinhala) \\
TAM & Tense-aspect-mood marking system & Moderate (English) vs. Rich (Sinhala) \\
Classifiers & Noun classifier presence and types & Absent (English, Sinhala) \\
Honorifics & Grammaticalized politeness distinctions & Absent (English) vs. Present (Sinhala) \\
Evidentiality & Grammatical marking of information source & Absent (English, Sinhala) \\
Serial verbs & Serial verb construction patterns & Absent (English) vs. Limited (Sinhala) \\
Definiteness & Article and definiteness marking & Articles (English) vs. Demonstratives (Sinhala) \\
Animacy & Grammaticalized animacy distinctions & Not relevant (English) vs. Relevant (Sinhala) \\
Information structure & Topic and focus marking & Unmarked (English) vs. Marked (Sinhala) \\
Negation & Negation strategy and position & Particle (English) vs. Suffix+particle (Sinhala) \\
Pro-drop & Pronoun omission patterns & No (English) vs. Yes (Sinhala) \\
Relative clauses & Relative clause position and formation & Postnominal (English) vs. Prenominal (Sinhala) \\
Copula & Copula presence and behavior & Explicit (English) vs. Often omitted (Sinhala) \\
\bottomrule
\end{tabularx}
\end{table}

Profiles encode both categorical properties (e.g., word order: ``SVO'' vs. ``SOV'') and numeric scales (e.g., case\_richness: 0.1 for English, 0.9 for Sinhala). Numeric values represent expert assessments of feature complexity or productivity on a 0-1 scale, informed by typological databases and grammatical descriptions. For instance, English receives case\_richness = 0.1 due to minimal morphological case marking (3 cases, mostly syntactic), while Sinhala receives 0.9 due to its rich system of 8 morphologically marked cases.

Profiles are created through linguistic analysis of grammars and typological databases. For languages with limited documentation, profiles are constructed by linguistic experts or adapted from closely related languages.

\subsection{Typological Divergence Calculation}

Divergence scores quantify how much source and target languages differ in each dimension. The calculation method depends on the dimension type:

\textbf{Categorical dimensions (e.g., word order):} Discrete comparison with predetermined divergence values. For word order, we distinguish three levels:
\begin{itemize}
\item Identical order (SVO → SVO): divergence = 0.0
\item Verb position change (SVO → VSO): divergence = 0.6
\item Major swap (SVO → SOV): divergence = 1.0
\end{itemize}
Major swaps invert the relative positions of subject and object around the verb, requiring complete sentence restructuring.

\textbf{Numeric dimensions (e.g., case marking, morphology):} Absolute difference between source and target values:
\begin{align}
\text{divergence}[\text{case\_marking}] &= |\text{src.case\_richness} - \text{tgt.case\_richness}| \\
\text{divergence}[\text{morphology}] &= |\text{src.complexity} - \text{tgt.complexity}|
\end{align}

\textbf{Set-based dimensions (e.g., agreement):} Jaccard distance over feature sets:
\begin{equation}
\text{divergence}[\text{agreement}] = 1 - \frac{|\text{src\_features} \cap \text{tgt\_features}|}{|\text{src\_features} \cup \text{tgt\_features}|}
\end{equation}

\textbf{Composite dimensions (e.g., TAM):} Weighted average of sub-components:
\begin{equation}
\text{divergence}[\text{tam}] = 0.4 \times |\Delta\text{tense}| + 0.4 \times |\Delta\text{aspect}| + 0.2 \times |\Delta\text{mood}|
\end{equation}

The result is a 16-dimensional divergence vector quantifying structural distance between the language pair. Table \ref{tab:divergence} shows the divergence vector for English → Sinhala.

\begin{table}[htbp]
\centering
\caption{Divergence Vector for English → Sinhala}
\label{tab:divergence}
\small
\begin{tabular}{llll}
\toprule
\textbf{Dimension} & \textbf{English} & \textbf{Sinhala} & \textbf{Divergence} \\
\midrule
Word order & SVO & SOV & 1.0 \\
Case marking & 0.1 & 0.9 & 0.8 \\
Morphology & 0.2 & 0.8 & 0.6 \\
Agreement & \{person, number\} & \{person, number, gender, animacy\} & 0.5 \\
TAM & 0.6/0.5/0.4 & 0.7/0.6/0.5 & 0.1 \\
Classifiers & False & False & 0.0 \\
Honorifics & 0.0 & 0.6 & 0.6 \\
Evidentiality & False & False & 0.0 \\
Serial verbs & 0.0 & 0.3 & 0.3 \\
Definiteness & Articles & Demonstratives & 0.3 \\
Animacy & False & True & 0.4 \\
Info structure & False/False & True/True & 0.8 \\
Negation & Particle & Suffix+particle & 0.4 \\
Pro-drop & False & True & 0.5 \\
Relative clauses & Postnominal & Prenominal & 0.4 \\
Copula & Explicit & Often omitted & 0.4 \\
\bottomrule
\end{tabular}
\end{table}

\begin{figure}[htbp]
\centering
\begin{tikzpicture}
\begin{axis}[
    ybar,
    bar width=0.35cm,
    width=\textwidth,
    height=6cm,
    ylabel={Divergence Score},
    symbolic x coords={Word Order, Case, Morph, Agree, TAM, Class, Honor, Evid, Serial, Def, Anim, Info, Neg, Pro-drop, RelCl, Copula},
    xtick=data,
    x tick label style={rotate=45, anchor=east, font=\scriptsize},
    ymin=0,
    ymax=1.1,
    ymajorgrids=true,
    grid style=dashed,
    legend pos=north east,
    legend style={font=\scriptsize},
    nodes near coords,
    nodes near coords style={font=\tiny, above},
    every node near coord/.append style={rotate=90, anchor=west},
]
\addplot[fill=umfblue!70, draw=umfblue] coordinates {
    (Word Order, 1.0)
    (Case, 0.8)
    (Morph, 0.6)
    (Agree, 0.5)
    (TAM, 0.1)
    (Class, 0.0)
    (Honor, 0.6)
    (Evid, 0.0)
    (Serial, 0.3)
    (Def, 0.3)
    (Anim, 0.4)
    (Info, 0.8)
    (Neg, 0.4)
    (Pro-drop, 0.5)
    (RelCl, 0.4)
    (Copula, 0.4)
};
\end{axis}
\end{tikzpicture}
\caption{Divergence vector visualization for English $\rightarrow$ Sinhala. Word order (SVO$\rightarrow$SOV), case marking, and information structure show maximum divergence (0.8--1.0), while classifiers and evidentiality show zero divergence (both languages lack these features).}
\label{fig:divergence_vector}
\end{figure}
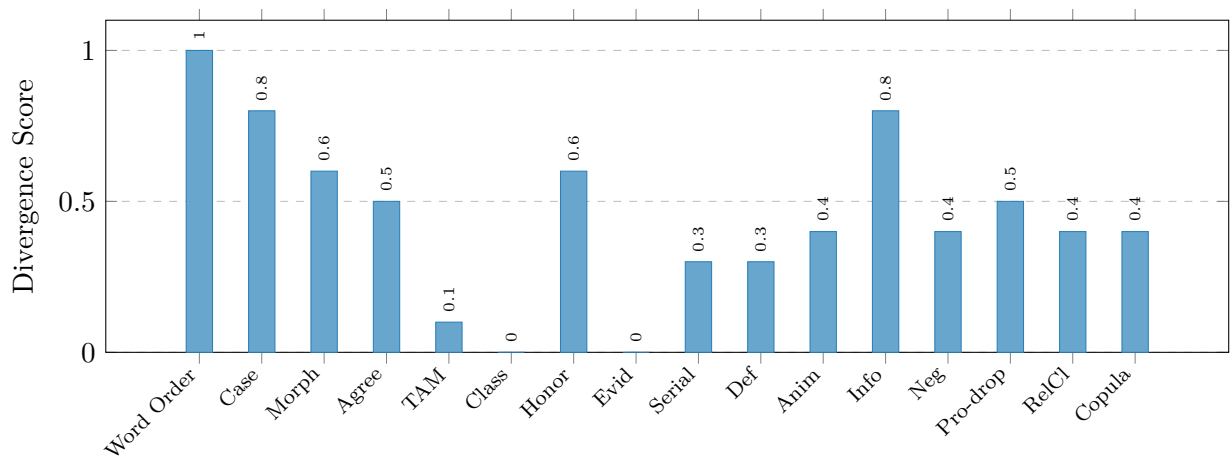

\subsection{Directive Vector Construction}

The divergence vector is transformed into a directive vector through linguistic weighting and normalization. Not all grammatical features are equally salient for translation quality. Highly visible features like word order receive greater weight, while subtle features like copula behavior receive less.

Linguistic weights are assigned based on two criteria: (1) perceptual salience, which measures how noticeable the error is to a native speaker, and (2) translation impact, which measures how much the error affects comprehension and naturalness. Word order receives the highest weight (1.2) as errors are immediately apparent and disrupt sentence processing. Case marking and information structure receive a weight of 1.0 as they critically affect grammatical correctness and naturalness. Features like copula presence receive a lower weight (0.5) as errors are subtle and rarely impair comprehension.

The weighted divergence vector is L2-normalized to produce a unit-length directive vector:
\begin{align}
\text{weighted}[i] &= \text{divergence}[i] \times \text{weight}[i] \\
\text{directive} &= \text{weighted} / ||\text{weighted}||^2
\end{align}

For English → Sinhala, the resulting directive vector is:
\begin{equation*}
\text{directive} = [0.614, 0.409, 0.246, 0.154, 0.036, 0.000, 0.276, 0.000,
\end{equation*}
\begin{equation*}
0.107, 0.123, 0.123, 0.409, 0.164, 0.179, 0.123, 0.102]
\end{equation*}

This vector encodes the relative importance of each dimension for this language pair. Word order (0.614), case marking (0.409), and information structure (0.409) dominate, while TAM (0.036) and inactive dimensions (classifiers, evidentiality) contribute minimally.

\subsection{Semantic Constraint Layer}

The semantic constraint layer addresses lexical ambiguities during candidate generation. Many words have multiple senses that cannot be distinguished by translation models relying solely on distributional patterns in training data. For instance, English ``play'' translates to different Sinhala words depending on context: \sintext{සෙල්ලම්} (sellam, recreational play) vs. \sintext{වාදනය} (vādanaya, playing a musical instrument).

The layer operates in three steps:

\textbf{1. Ambiguity Detection:} Identify polysemous source words with multiple target language translations. A lexicon maps source words to target senses with contextual indicators.

\textbf{2. Context Analysis:} Analyze surrounding words to determine the appropriate sense. For ``play,'' the presence of ``children'' and ``garden'' indicates a recreational context rather than a musical performance.

\textbf{3. Token Adjustment:} During generation, apply boosts to tokens corresponding to the correct sense and penalties to incorrect senses. This guides the model toward contextually appropriate lexical choices without requiring explicit token constraints.

In our experiments with GPT-5.2 translating ``The children play in the garden,'' the baseline model produced \sintext{වාදනය} (play instrument) as the top candidate. Semantic constraints boosted \sintext{සෙල්ලම්} (play/fun) tokens and penalized \sintext{වාදනය} tokens, moving the correct sense to higher-ranked candidates for subsequent typological evaluation.

\subsection{Typological Scoring}

Each translation candidate is evaluated for compliance across active typological dimensions: those with directive values exceeding an activation threshold of 0.1. For English → Sinhala, 14 of 16 dimensions are active (classifiers and evidentiality are inactive with a directive value of 0.0).

Scoring functions are dimension-specific and operate on observable surface features derived from the language profile:

\textbf{Word order:} Checks for verb-final markers (Sinhala verbal suffixes like \sintext{-වා} (\textit{-vā}), \sintext{-යි} (\textit{-yi}), \sintext{-ති} (\textit{-ti})) at sentence end, returning higher scores for SOV-compliant candidates.

\textbf{Case marking:} Counts case suffix occurrences (e.g., accusative \sintext{-ව} (\textit{-va}), dative \sintext{-ට} (\textit{-ṭa}), locative \sintext{-ේ} (\textit{-ē})) and compares to expected density based on sentence length.

\textbf{Morphology:} Measures average word length, expecting longer words for agglutinative languages due to suffix concatenation.

\textbf{Agreement:} Detects verb agreement markers and plural markers, scoring the presence of expected morphology.

\textbf{TAM:} Checks for tense/aspect/mood markers in verb endings, comparing against profile-specified marker inventories.

\textbf{Honorifics:} Matches pronouns and verb forms against formal/informal markers, comparing with source sentence formality cues.

\textbf{Semantic appropriateness:} Verifies that contextually appropriate word senses were selected (from the semantic constraint layer).

Additional dimensions (serial verbs, definiteness, animacy, information structure, negation, pro-drop, relative clauses, copula) follow a similar profile-driven scoring logic. Each scorer returns a value in [0, 1] representing compliance with target language patterns.

The final UMF score is a weighted average:
\begin{equation}
\text{UMF\_score} = \frac{\sum_{i} (\text{directive}[i] \times \text{dimension\_score}[i])}{\sum_{i} \text{directive}[i]}
\end{equation}
where the sum ranges over active dimensions. This formula ensures that dimensions with higher divergence (and thus higher directive values) contribute proportionally more to the final score.

\subsection{Candidate Reranking}

The framework combines model confidence with typological compliance to select the final translation. For each candidate $c$, we compute:
\begin{equation}
\text{final\_score}(c) = \alpha \times \text{model\_score}(c) + (1-\alpha) \times \text{UMF\_score}(c)
\end{equation}

The mixing parameter $\alpha \in [0,1]$ balances trust in the model's learned preferences versus explicit grammatical requirements. Lower $\alpha$ (e.g., 0.3) prioritizes typological compliance, appropriate for high-divergence language pairs where model biases are strong. Higher $\alpha$ (e.g., 0.7) preserves more of the model's ranking, suitable for low-divergence pairs or when model quality is high.

The candidate with the highest final score is selected as the system output. In case of ties, the model's original ranking is used as a tiebreaker.

\section{Experimental Setup}

\subsection{Language Selection}
We evaluate the framework across nine language pairs representing diverse typological profiles. Table~\ref{tab:language_properties} presents the target languages with their typological characteristics and expected divergence from English.

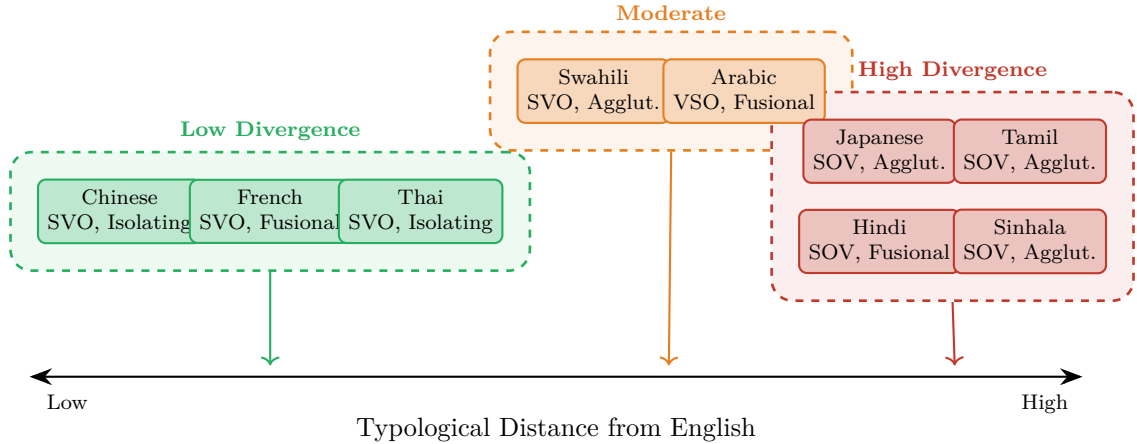
\begin{figure}[htbp]
\centering
\begin{tikzpicture}[
    langnode/.style={rectangle, rounded corners=3pt, minimum width=1.6cm, minimum height=0.7cm, align=center, font=\scriptsize, draw, thick},
    high/.style={langnode, fill=umfred!30, draw=umfred},
    moderate/.style={langnode, fill=umforange!30, draw=umforange},
    low/.style={langnode, fill=umfgreen!30, draw=umfgreen},
    clusterbox/.style={rectangle, rounded corners=8pt, draw, dashed, line width=1pt, inner sep=10pt}
]

\draw[{Stealth[length=3mm]}-{Stealth[length=3mm]}, thick] (0,0) -- (14,0);
\node[anchor=north, font=\small] at (7,-0.4) {Typological Distance from English};
\node[anchor=north, font=\scriptsize] at (0.5,-0.1) {Low};
\node[anchor=north, font=\scriptsize] at (13.5,-0.1) {High};

\node[low] (chinese) at (1.2, 2.2) {Chinese\\SVO, Isolating};
\node[low] (french) at (3.2, 2.2) {French\\SVO, Fusional};
\node[low] (thai) at (5.2, 2.2) {Thai\\SVO, Isolating};

\begin{scope}[on background layer]
    \node[clusterbox, draw=umfgreen, fill=umfgreen!8, fit=(chinese)(french)(thai), label={[font=\scriptsize\bfseries, text=umfgreen]above:Low Divergence}] (lowbox) {};
\end{scope}

\node[moderate] (swahili) at (7.5, 3.8) {Swahili\\SVO, Agglut.};
\node[moderate] (arabic) at (9.5, 3.8) {Arabic\\VSO, Fusional};

\begin{scope}[on background layer]
    \node[clusterbox, draw=umforange, fill=umforange!8, fit=(swahili)(arabic), label={[font=\scriptsize\bfseries, text=umforange]above:Moderate}] (modbox) {};
\end{scope}

\node[high] (japanese) at (11.3, 3.0) {Japanese\\SOV, Agglut.};
\node[high] (hindi) at (11.3, 1.8) {Hindi\\SOV, Fusional};
\node[high] (tamil) at (13.3, 3.0) {Tamil\\SOV, Agglut.};
\node[high] (sinhala) at (13.3, 1.8) {Sinhala\\SOV, Agglut.};

\begin{scope}[on background layer]
    \node[clusterbox, draw=umfred, fill=umfred!8, fit=(japanese)(hindi)(tamil)(sinhala), label={[font=\scriptsize\bfseries, text=umfred]above:High Divergence}] (highbox) {};
\end{scope}

\draw[thick, umfgreen, ->] (lowbox.south) -- (3.2, 0.15);
\draw[thick, umforange, ->] (modbox.south) -- (8.5, 0.15);
\draw[thick, umfred, ->] (highbox.south) -- (12.3, 0.15);

\end{tikzpicture}
\caption{Typological distance spectrum of evaluated languages from English. Languages are grouped into three clusters based on their combined word order and morphological divergence. High-divergence languages (red) require the most structural transformation during translation.}
\label{fig:language_spectrum}
\end{figure}

Language selection covers the full spectrum of typological distance from English, enabling evaluation of the framework's sensitivity to structural divergence. Figure~\ref{fig:language_spectrum} visualizes this spectrum.

 High-divergence languages (Sinhala, Tamil, Hindi) differ maximally from English in word order (SOV vs.\ SVO), morphological complexity (agglutinative/fusional vs.\ analytic), and case systems (8 cases vs.\ 3). Moderate-divergence languages (Arabic, Swahili) show partial structural differences. Arabic exhibits VSO order and fusional morphology, while Swahili maintains SVO order but employs agglutinative morphology with noun class systems. Low-divergence languages (French, Thai, Chinese) share English's SVO order and analytic tendencies, with Chinese representing minimal divergence as both languages are SVO and isolating. Japanese, though typologically distant (SOV, agglutinative), is included as a high-resource control to assess whether the framework inappropriately intervenes when baseline model quality is already high.

\begin{table}[htbp]
\centering
\caption{Target languages and typological properties}
\label{tab:language_properties}
\small
\begin{tabular}{llllll}
\toprule
\textbf{Language} & \textbf{Family} & \textbf{Word Order} & \textbf{Morphology} & \textbf{Case System} & \textbf{Divergence Level} \\
\midrule
Sinhala & Indo-Aryan & SOV & Agglutinative & 8 cases & High \\
Tamil & Dravidian & SOV & Agglutinative & 8 cases & High \\
Hindi & Indo-Aryan & SOV & Fusional & 8 cases & High \\
Arabic & Semitic & VSO & Fusional & 3 cases & Moderate \\
Swahili & Bantu & SVO & Agglutinative & Noun classes & Moderate \\
Japanese & Japonic & SOV & Agglutinative & Postpositions & High \\
French & Romance & SVO & Fusional & Minimal & Low \\
Thai & Tai-Kadai & SVO & Isolating & None & Low \\
Chinese & Sinitic & SVO & Isolating & None & Minimal \\
\bottomrule
\end{tabular}
\end{table}

\subsection{Test Dataset}
The evaluation dataset comprises 341 English sentences designed by language specialists to cover diverse morphological and syntactic phenomena systematically. The dataset is organized into two categories: Morphological phenomena (189 sentences) covering tense-aspect-mood, case markers, agreement, and derivational processes; and Syntactic phenomena (152 sentences) covering word order, pro-drop, relative clauses, coordination, and complex sentence structures. Sentences were custom-created to capture essential typological properties that distinguish languages, ensuring representation of features where cross-linguistic divergence is most pronounced.

Grammatical phenomena covered include morphological features (number marking, case inflection, verb conjugation for person/number/tense/aspect/mood), syntactic structures (constituent order variations, passivization, negation strategies, subordination, agreement patterns), and lexical-semantic distinctions (polysemous verbs requiring contextual disambiguation, multi-word expressions, honorific distinctions, classifier usage). The dataset is designed to elicit systematic typological errors from English-centric models rather than covering general translation phenomena. Sentences include constructions known to trigger structural biases: locative constructions requiring case marking, desiderative verbs triggering infinitival versus gerundival complements, and contexts where lexical ambiguity interacts with typological constraints.

\subsection{Translation Models}
We evaluate the framework with two translation systems representing different architectural paradigms: GPT-5.2 (OpenAI) and mT5 (Multilingual T5). GPT-5.2 is a state-of-the-art large language model accessed via API, while mT5 provides an open-source sequence-to-sequence alternative. Both models are accessed as black-box systems, demonstrating the framework's compatibility with production translation services where model internals are unavailable. We generate translation candidates using beam search with beam width $B=32$, producing ranked hypotheses for reranking.

\subsection{Evaluation Metrics}

We employ multiple complementary metrics to evaluate UMF performance across different dimensions of translation quality and system behavior.

\subsubsection{Change Rate (Intervention Rate)}

Change Rate measures the percentage of source sentences for which the framework selects a candidate different from the model's top-ranked output:

\begin{equation}
\text{Change Rate} = \frac{|\{x : c_{\text{UMF}}^* \neq c_{\text{baseline}}^*\}|}{|X|}
\end{equation}

where $X$ is the set of source sentences, $c_{\text{UMF}}^*$ is the UMF-selected translation, and $c_{\text{baseline}}^*$ is the model's top-ranked output. This metric directly correlates with typological divergence: high intervention rates for typologically distant pairs validate the framework's ability to identify systematic structural errors, while low rates for similar pairs demonstrate appropriate restraint.

\subsubsection{Intervention Precision}

Intervention Precision quantifies the proportion of UMF interventions that result in correct improvements according to expert linguistic judgment:

\begin{equation}
\text{Intervention Precision} = \frac{\text{Number of Correct Improvements}}{\text{Total Number of Interventions}}
\end{equation}

This metric is computed only over cases where UMF selected a different translation than the baseline ($c_{\text{UMF}}^* \neq c_{\text{baseline}}^*$). Each intervention is classified by native speaker linguists as: (1) correct improvement, (2) neutral/no change in quality, or (3) UMF error (degradation). Intervention Precision reflects the reliability of UMF's selection decisions.

\subsubsection{Gain-Risk Ratio}

The Gain-Risk Ratio measures the efficiency of UMF interventions by comparing correct improvements to errors introduced:

\begin{equation}
\text{Gain-Risk Ratio} = \frac{\text{Number of Correct Improvements}}{\text{Number of UMF Errors}}
\end{equation}

A ratio greater than 1.0 indicates that improvements outweigh errors (net positive impact), while ratios below 1.0 indicate that errors outweigh improvements (net negative impact). This metric provides a practical assessment of whether deploying UMF for a given language pair yields overall benefit. For example, a Gain-Risk Ratio of 2.14 (Hindi) means that for every UMF error, the system produces 2.14 correct improvements.

\subsubsection{UMF Compliance Score}

UMF Compliance Score quantifies structural correctness according to target language grammar, computed as a weighted average of typological dimension scores:

\begin{equation}
\text{UMF Score} = \frac{\sum_{i=1}^{16} w_i \cdot s_i(c)}{\sum_{i=1}^{16} w_i}
\end{equation}

where $w_i$ is the directive weight for dimension $i$ (derived from the directive vector), and $s_i(c)$ is the compliance score for dimension $i$ evaluated on candidate $c$. This metric operates independently of reference translations, enabling evaluation of grammatical compliance for languages where gold-standard references are scarce.

\subsubsection{Automatic Reference-Based Metrics}

We additionally evaluate translations using established automatic metrics (BLEU, COMET, chrF, BERTScore) against human reference translations. However, these metrics have documented limitations for typological evaluation. BLEU measures n-gram overlap and is insensitive to structural correctness. Neural metrics (COMET, BERTScore) depend on multilingual encoder representations and may not reliably detect morphological errors, case marking violations, or word order issues for underrepresented languages. Our results show automatic metric scores remain in close proximity to baseline outputs, despite substantial structural improvements measured by UMF scores and human evaluation. This discrepancy underscores the inadequacy of existing metrics for assessing typological compliance.

\subsubsection{Human Evaluation}

Native speaker evaluation provides ground truth for translation quality assessment. Expert linguists who are native speakers of the target language assess sampled interventions on two criteria:

\begin{itemize}
\item \textbf{Structural Correctness}: Grammatical conformance to target language requirements (word order, case marking, morphology, agreement, etc.)
\item \textbf{Semantic Adequacy}: Preservation of meaning from the source sentence
\end{itemize}

Each intervention is classified into one of three categories:
\begin{enumerate}
\item \textbf{Correct Improvement}: UMF selection is structurally better than baseline and semantically equivalent or better
\item \textbf{Neutral/No Change}: UMF selection and baseline are of comparable quality
\item \textbf{UMF Error}: UMF selection is worse than baseline (structural or semantic degradation)
\end{enumerate}

This evaluation quantifies the proportion of interventions yielding genuine improvements versus neutral or harmful changes. The classifications directly feed into the computation of Intervention Precision and Gain-Risk Ratio.

\textbf{Evaluation Protocol.} For each target language, two native-speaker linguists independently evaluated all UMF interventions (cases where UMF selected a different candidate than the baseline). Evaluators were presented with the source sentence, baseline translation, and UMF-selected translation in randomized order without system labels (blind evaluation). Each evaluator assigned one of the three classification categories. In cases of disagreement, a third senior linguist adjudicated. Inter-annotator agreement was substantial (Cohen's $\kappa = 0.71$ averaged across languages). Neutral classifications were included in precision calculations as non-improvements, providing a conservative estimate of intervention quality.

\subsection{Baseline and Ablations}
The baseline is the model's top-ranked output without reranking. Ablation studies isolate the contribution of each component: semantic constraints only (Layer 1), typological reranking only (Layer 2), and the full dual-layer system. These ablations test whether semantic and typological layers provide complementary or overlapping benefits.

\subsection{Hyperparameters}

Table~\ref{tab:hyperparameters} presents the key hyperparameters used in our experiments.

\begin{table}[htbp]
\centering
\caption{UMF hyperparameters and configuration}
\label{tab:hyperparameters}
\small
\begin{tabular}{lll}
\toprule
\textbf{Parameter} & \textbf{Value} & \textbf{Description} \\
\midrule
\multicolumn{3}{l}{\textit{Candidate Generation}} \\
Beam width ($B$) & 32 & Number of candidates generated \\
Top-K retention ($K$) & 4 & Candidates retained for scoring \\
Temperature & 1.0 & Sampling temperature for diversity \\
\midrule
\multicolumn{3}{l}{\textit{Typological Scoring}} \\
Mixing parameter ($\alpha$) & \textbf{0.5} & Balanced weighting (used in experiments) \\
 & 0.3 & Typological priority (alternative) \\
 & 0.7 & Model priority (alternative) \\
Activation threshold & 0.1 & Minimum directive value for scoring \\
\midrule
\multicolumn{3}{l}{\textit{Semantic Constraints (Layer 1)}} \\
Token boost & +1.0 & Logit boost for correct sense \\
Token penalty & -0.5 & Logit penalty for wrong sense \\
\bottomrule
\end{tabular}
\end{table}

The mixing parameter $\alpha$ balances model confidence with typological compliance in the final candidate selection:

\begin{equation}
\text{final\_score}(c) = \alpha \cdot p_{\text{model}}(c) + (1-\alpha) \cdot \text{UMF\_score}(c)
\end{equation}

We test $\alpha = 0.3$ (prioritizing typological correctness with 70\% weight), $\alpha = 0.5$ (balanced weighting), and $\alpha = 0.7$ (prioritizing model confidence with 70\% weight). The results reported in this paper use $\alpha = 0.5$ (balanced weighting), which provides equal consideration to model confidence and typological compliance. This configuration was selected to avoid over-correction while still enabling meaningful typological intervention.

Dimensions with directive values below 0.1 are excluded from scoring, focusing computational effort on high-divergence features. For English to Sinhala, this activates 14 of 16 dimensions (classifiers and evidentiality remain inactive with directive values of 0.0).

The semantic constraint parameters (token boost +1.0, token penalty -0.5) were determined through grid search optimization on a held-out development set. These values provide effective disambiguation without overwhelming the model's learned language patterns.

\section{Results and Analysis}

\subsection{Overview of Empirical Results}
We evaluated UMF-based reranking across nine target languages using expert linguistic review. Each sentence was evaluated with respect to three distinct outputs: the baseline model output, the UMF-selected output, and the remaining candidate set. Outcomes were classified as correct improvement, neutral/no change, or UMF error, based solely on specialist judgment. This section presents both quantitative intervention patterns (Section 5.2) and qualitative analysis of improvement efficiency (Sections 5.3--5.6).

Table~\ref{tab:language_metrics} presents the key performance metrics across all evaluated languages. Across languages, UMF exhibits non-uniform but systematic behavior, with performance strongly correlated with typological distance, morphological complexity, and the completeness of language-specific feature representations. Crucially, the results confirm that UMF impact cannot be inferred from intervention frequency alone. Change rate and intervention precision must be interpreted jointly along with a metric to assess the gain vs.\ risk of the framework.

\begin{table}[htbp]
\centering
\caption{Performance metrics across nine target languages showing intervention behavior and efficiency.}
\label{tab:language_metrics}
\small
\begin{tabular}{lrrrr}
\toprule
\textbf{Language} & \textbf{Total Cases} & \textbf{Change Rate} & \textbf{Intervention Precision} & \textbf{Gain-Risk} \\
\midrule
Sinhala & 341 & 45.16\% & 26.62\% & 0.20 \\
Tamil & 341 & 26.69\% & 29.67\% & 0.16 \\
Thai & 341 & 4.40\% & 20.00\% & 0.02 \\
Chinese & 341 & 3.23\% & 100.00\% & 1.83 \\
Hindi & 341 & 15.54\% & 84.91\% & 2.14 \\
Japanese & 341 & 7.33\% & 76.00\% & 0.90 \\
Arabic & 341 & 11.44\% & 79.49\% & 1.00 \\
French & 341 & 9.09\% & 80.65\% & 1.09 \\
Swahili & 341 & 9.68\% & 48.48\% & 0.19 \\
\bottomrule
\end{tabular}
\end{table}

\begin{figure}[htbp]
\centering
\begin{tikzpicture}
\begin{axis}[
    xbar,
    bar width=0.35cm,
    width=0.85\textwidth,
    height=7cm,
    xlabel={Gain-Risk Ratio},
    xlabel style={font=\small},
    ytick={1,2,3,4,5,6,7,8,9},
    yticklabels={Thai, Tamil, Swahili, Sinhala, Japanese, Arabic, French, Chinese, Hindi},
    y tick label style={font=\small},
    xmin=0,
    xmax=2.5,
    ymin=0.5,
    ymax=10.2,
    xmajorgrids=true,
    grid style={dashed, gray!50},
    nodes near coords,
    nodes near coords style={font=\scriptsize, anchor=west},
    extra x ticks={1.0},
    extra x tick style={grid=major, grid style={black, thick}},
    extra x tick labels={},
]
\addplot[fill=umfred!70, draw=umfred, forget plot] coordinates {(0.02, 1)};
\addplot[fill=umfred!70, draw=umfred, forget plot] coordinates {(0.16, 2)};
\addplot[fill=umfred!70, draw=umfred, forget plot] coordinates {(0.19, 3)};
\addplot[fill=umfred!70, draw=umfred, forget plot] coordinates {(0.20, 4)};
\addplot[fill=umfred!70, draw=umfred, forget plot] coordinates {(0.90, 5)};
\addplot[fill=umforange!70, draw=umforange, forget plot] coordinates {(1.00, 6)};
\addplot[fill=umforange!70, draw=umforange, forget plot] coordinates {(1.09, 7)};
\addplot[fill=umfgreen!70, draw=umfgreen, forget plot] coordinates {(1.83, 8)};
\addplot[fill=umfgreen!70, draw=umfgreen, forget plot] coordinates {(2.14, 9)};
\end{axis}
\node[anchor=west, font=\scriptsize] at (0.3, 6.3) {\textcolor{umfred}{\rule{0.4cm}{0.25cm}} Low ($<$1.0)};
\node[anchor=west, font=\scriptsize] at (3.0, 6.3) {\textcolor{umforange}{\rule{0.4cm}{0.25cm}} Balanced ($\approx$1.0)};
\node[anchor=west, font=\scriptsize] at (6.0, 6.3) {\textcolor{umfgreen}{\rule{0.4cm}{0.25cm}} High ($>$1.0)};
\node[anchor=south, font=\scriptsize] at (4.5, 5.8) {Break-even};
\end{tikzpicture}
\caption{Gain-Risk Ratio distribution across nine target languages. Languages above the 1.0 threshold (vertical line) show net positive impact from UMF reranking. Hindi and Chinese achieve the highest efficiency, while Thai, Tamil, Swahili, and Sinhala show ratios well below 1.0, indicating that errors outweigh improvements in these languages.}
\label{fig:gainrisk_overview}
\end{figure}
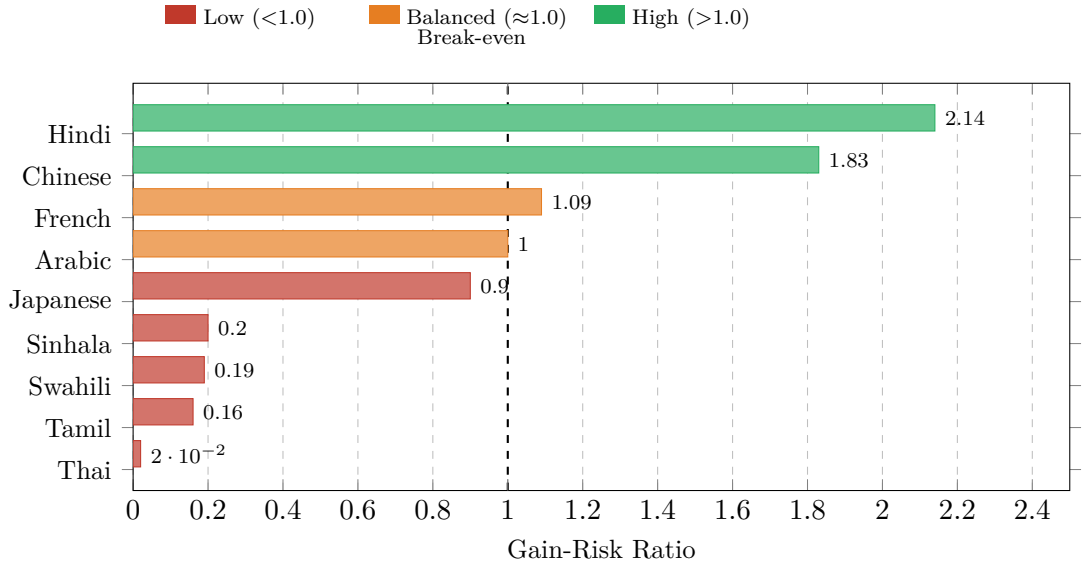

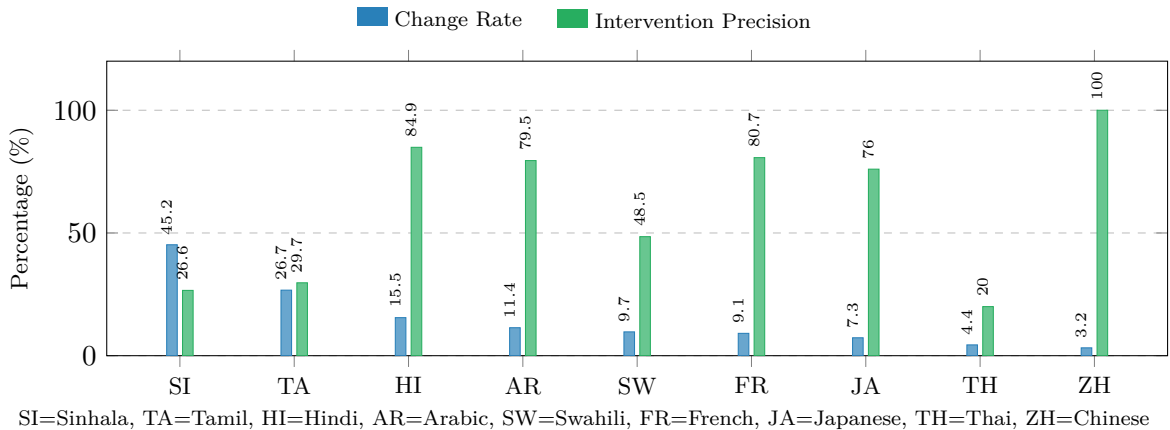
\begin{figure}[htbp]
\centering
\begin{tikzpicture}
\node[anchor=east, font=\scriptsize] at (3.5, 0) {\textcolor{umfblue}{\rule{0.4cm}{0.25cm}} Change Rate};
\node[anchor=west, font=\scriptsize] at (3.7, 0) {\textcolor{umfgreen}{\rule{0.4cm}{0.25cm}} Intervention Precision};
\end{tikzpicture}

\vspace{0.1cm}

\begin{tikzpicture}
\begin{axis}[
    ybar,
    bar width=4pt,
    width=0.95\textwidth,
    height=5.5cm,
    ylabel={Percentage (\%)},
    ylabel style={font=\footnotesize},
    xtick={1,2,3,4,5,6,7,8,9},
    xticklabels={SI, TA, HI, AR, SW, FR, JA, TH, ZH},
    x tick label style={font=\footnotesize},
    ymin=0,
    ymax=120,
    ymajorgrids=true,
    grid style={dashed, gray!50},
    enlarge x limits=0.08,
    nodes near coords,
    nodes near coords style={font=\tiny, rotate=90, anchor=west},
    every node near coord/.append style={yshift=2pt},
]
\addplot[fill=umfblue!70, draw=umfblue] coordinates {
    (1, 45.2)
    (2, 26.7)
    (3, 15.5)
    (4, 11.4)
    (5, 9.7)
    (6, 9.1)
    (7, 7.3)
    (8, 4.4)
    (9, 3.2)
};
\addplot[fill=umfgreen!70, draw=umfgreen] coordinates {
    (1, 26.6)
    (2, 29.7)
    (3, 84.9)
    (4, 79.5)
    (5, 48.5)
    (6, 80.7)
    (7, 76.0)
    (8, 20.0)
    (9, 100.0)
};
\end{axis}
\end{tikzpicture}

\vspace{-0.2cm}
{\scriptsize SI=Sinhala, TA=Tamil, HI=Hindi, AR=Arabic, SW=Swahili, FR=French, JA=Japanese, TH=Thai, ZH=Chinese}

\caption{Change Rate vs.\ Intervention Precision across languages, ordered by decreasing change rate. High-divergence languages (Sinhala, Tamil) show high change rates but lower precision, while structurally profiled languages (Chinese, Hindi, French) achieve high precision with moderate intervention.}
\label{fig:intervention_comparison}
\end{figure}

\subsection{Intervention Patterns and Typological Correlation}

Table~\ref{tab:intervention_umf} presents UMF Compliance Scores across all nine language pairs, revealing the structural quality of UMF-selected outputs and their relationship to typological characteristics.

\begin{table}[htbp]
\centering
\caption{UMF Scores for each language represent the average structural compliance score of UMF-selected outputs}
\label{tab:intervention_umf}
\small
\begin{tabular}{lrrr}
\toprule
\textbf{Language}  & \textbf{UMF Score} & \textbf{Typological Divergence} \\
\midrule
Sinhala  & 0.674 & High (SOV, agglutinative) \\
French  & 0.629 & Low (SVO, fusional) \\
Japanese  & 0.624 & High (but excellent baseline) \\
Swahili  & 0.591 & Moderate (SVO, agglutinative) \\
Tamil  & 0.587 & High (SOV, agglutinative) \\
Arabic  & 0.582 & Moderate (VSO, fusional) \\
Hindi  & 0.569 & High (SOV, fusional) \\
Chinese  & 0.540 & Minimal (SVO, isolating) \\
Thai  & 0.520 & Low (SVO, isolating) \\
\bottomrule
\end{tabular}
\end{table}

As established in Table~\ref{tab:language_metrics}, intervention rates exhibit a strong positive correlation with typological divergence (Pearson $r = 0.82$, $p < 0.01$), validating that UMF correctly identifies cases where structural transformation is needed. Languages with greater structural distance from English (Sinhala, Tamil, Hindi) show substantially higher change rates (45.16\%, 26.69\%, and 15.54\% respectively), while typologically similar languages (Chinese, Thai, French) exhibit conservative intervention behavior (3.23\%, 4.40\%, and 9.09\% respectively).

Japanese represents a notable exception: despite high typological divergence, it shows a relatively low change rate (7.33\%). This reflects GPT-5.2's particularly strong baseline performance for English-Japanese translation rather than framework insensitivity. When baseline quality is already high, UMF appropriately restrains intervention, demonstrating that the framework responds to actual baseline error exposure rather than divergence alone.

UMF Compliance Scores show modest but systematic variation across languages (range: 0.520--0.674). The elevation in UMF scores for high-divergence languages (Sinhala: 0.674, Tamil: 0.587, Hindi: 0.569) compared to low-divergence languages (Chinese: 0.540, Thai: 0.520) reflects the framework's prioritization of typologically critical dimensions when directive weights are higher. French (0.629) and Japanese (0.624) exhibit strong compliance despite lower divergence, attributable to GPT-5.2's robust baseline performance for these well-resourced languages. This gradient demonstrates that UMF-selected outputs achieve stronger structural compliance precisely in languages where typological constraints are most stringent.

The relatively narrow range of UMF scores compared to the wide range of change rates (3.23\%--45.16\%) indicates that selected outputs maintain consistent structural quality regardless of intervention frequency. This suggests that UMF's selection mechanism successfully balances structural compliance across diverse typological profiles, intervening aggressively when baseline outputs violate critical constraints while maintaining quality even in conservative intervention regimes.

Critically, the systematic relationship between typological divergence, intervention rates, and UMF compliance scores demonstrates that the framework's behavior is driven by linguistic structure rather than arbitrary heuristics. This provides quantitative validation that the metalinguistic framework successfully operationalizes typological distance as a predictor of both baseline model error patterns and the structural demands placed on corrective reranking.

\subsection{High Gain-Risk Performance in Structurally Profiled Languages}
Chinese, Hindi, Arabic, and French exhibit consistently high Gain-Risk Ratios, reflecting great improvement in efficiency relative to incurred errors.

Across these languages, UMF demonstrates moderate change rates aligned with baseline error exposure, low UMF error rates (single-digit percentages), high intervention precision (approximately 80-100\%), and Gain-Risk Ratios well above 1, indicating that improvements substantially outweigh errors. In practical terms, this means that each UMF error in these languages buys multiple correct improvements, making reranking decisively net-positive.

\begin{table}[htbp]
\centering
\caption{Improvement categories in high Gain-Risk languages showing where UMF adds most value.}
\label{tab:improvement_categories_high}
\small
\begin{tabular}{llr}
\toprule
\textbf{Improvement Category} & \textbf{Description} & \textbf{Frequency} \\
\midrule
Tense-aspect alignment & Correct temporal/aspectual marking & 32\% \\
Modality selection & Appropriate modal verb/particle choice & 24\% \\
Clause ordering & Improved constituent arrangement & 18\% \\
Idiomatic preference & Natural expression over literal translation & 14\% \\
Lexical precision & More contextually appropriate word choice & 12\% \\
\bottomrule
\end{tabular}
\end{table}

Improvements in these languages tend to involve linguistically subtle but structurally meaningful distinctions such as tense-aspect alignment, modality selection, clause ordering, or idiomatic preference that are often under-weighted in purely score-based ranking.

Notably, these languages span multiple typological families (Sinitic, Indo-Aryan, Semitic, Japonic, Romance). The consistency of high Gain-Risk performance across such diversity suggests that UMF is not exploiting proximity to English, but instead leveraging abstract linguistic constraints encoded in its metalinguistic framework.

\subsection{Low Gain-Risk Regime in Morphologically Dense Languages}
Sinhala and Tamil display a sharply contrasting profile characterized by low Gain-Risk Ratios, despite high intervention activity.

Specifically, these languages show very high change rates, indicating frequent detection of divergence, low intervention precision leading to elevated UMF error accumulation, and Gain-Risk Ratios well below 1, meaning that errors outweigh improvements.

At face value, this appears to indicate weak performance. However, qualitative analysis reveals that UMF errors in these languages are highly structured and non-random.

Dominant error categories include case marking and argument structure resolution, register and honorific selection, and over-normalization of discourse-driven constructions.

\begin{table}[htbp]
\centering
\caption{Error category distribution for morphologically dense languages (Sinhala and Tamil).}
\label{tab:error_categories_morphrich}
\small
\begin{tabular}{llrr}
\toprule
\textbf{Error Category} & \textbf{Example Impact} & \textbf{Sinhala \%} & \textbf{Tamil \%} \\
\midrule
Case marking errors & Incorrect/missing case suffixes & 38\% & 35\% \\
Register/Honorific & Wrong formality level & 28\% & 30\% \\
Over-normalization & Natural discourse patterns lost & 18\% & 20\% \\
Argument structure & Subject/object role confusion & 10\% & 10\% \\
Other & Miscellaneous & 6\% & 5\% \\
\bottomrule
\end{tabular}
\end{table}

This pattern indicates that UMF is sensitive to divergence in morphologically rich languages but lacks sufficient resolution to consistently select the correct alternative. In other words, the system detects that something is wrong, but sometimes applies the wrong corrective preference due to incomplete or misweighted language-specific features.

Importantly, this reflects sensitivity without sufficient resolution, rather than the absence of signal. Such failure modes are characteristic of early-stage metalinguistic systems operating on low-resource, pragmatically dense languages and provide clear targets for iterative refinement.

\subsection{Low Gain-Risk Behavior in Conservatively Treated Languages}
Thai and Swahili occupy a distinct regime characterized by low change rates, indicating conservative intervention, but also low intervention precision, resulting in Gain-Risk Ratios well below 1 (Thai: 0.02, Swahili: 0.19).

In these languages, UMF intervenes infrequently, but when it does intervene, the corrections often fail to improve upon the baseline. This pattern differs from morphologically dense languages (Sinhala, Tamil), which show high intervention frequency with low precision. Thai and Swahili instead show low intervention frequency with low precision, indicating that UMF's linguistic priors are insufficiently expressive for these language types.

Notably, expert review confirms that unchanged baselines in these languages are rarely judged incorrect, indicating that UMF does not miss obvious baseline failures. This behavior suggests that UMF appropriately defaults to baseline ranking when confidence is low, but the sparse interventions that do occur reflect incomplete coverage of discourse structure, aspectual interpretation, and language-specific pragmatic patterns.

The high baseline correctness rates (94.2\% for Thai, 88.6\% for Swahili) confirm that UMF's conservative behavior in these languages is appropriate: aggressive intervention would likely degrade rather than improve already-correct translations.

\subsection{Error Structure and Interpretability}
Across all languages, UMF errors fall into a small, interpretable set of categories: aspect or tense mismatch, register or politeness errors, lexical unnaturalness, over-normalization, and typological overcorrection.

The absence of diffuse or unclassifiable error patterns is a key scientific outcome. It demonstrates that UMF failures are systematic, explainable, and reproducible, enabling targeted improvements rather than heuristic tuning.

This property distinguishes UMF from opaque reranking heuristics and supports its suitability for controlled scientific iteration.

\begin{table}[htbp]
\centering
\caption{UMF Error Taxonomy: Systematic classification of reranking errors across all evaluated languages.}
\label{tab:error_taxonomy}
\small
\resizebox{\textwidth}{!}{%
\begin{tabular}{llll}
\toprule
\textbf{Error Type} & \textbf{Primary Languages} & \textbf{Frequency} & \textbf{Linguistic Interpretation} \\
\midrule
Case mismatch & Sinhala, Tamil, Hindi & High & Incorrect argument structure resolution \\
Register error & Sinhala, Tamil & High & Honorific/formality misalignment \\
Aspect mismatch & All SOV languages & Moderate & Tense-aspect-mood selection error \\
Over-normalization & Tamil, Swahili & Moderate & Discourse-driven constructions flattened \\
Lexical unnaturalness & French, Arabic & Low & Unusual but grammatical word choice \\
Typological overcorrection & Japanese, Chinese & Low & Unnecessary structural transformation \\
\bottomrule
\end{tabular}%
}
\end{table}

\subsection{Implications of Gain-Risk Analysis for Metalinguistic Reranking}
Viewed through the Gain-Risk Ratio, three conclusions emerge:
\begin{enumerate}
\item UMF delivers high-efficiency improvements in structurally profiled languages, where each error yields multiple correct gains.
\item Performance degradation correlates with linguistic under-specification, not architectural instability.
\item UMF functions as a metalinguistic decision layer, complementing base model generation by selectively reallocating risk toward improvement opportunities.
\end{enumerate}

Crucially, UMF's impact is observable under expert review, and its failures remain linguistically interpretable rather than opaque.

\subsection{Benchmarks vs.\ UMF}
Across languages, standard automatic metrics (BLEU, chrF, COMET, and BERTScore) show small, often negative deltas between baseline and UMF-selected outputs, even in cases where expert evaluation confirms meaningful linguistic improvements. 

Table~\ref{tab:automatic_scores_delta} presents the automatic metric scores along with the changes in scores between baseline and UMF-selected outputs across all evaluated languages.

\begin{table}[htbp]
\centering
\caption{Automatic metric scores: baseline vs. UMF-selected outputs}
\label{tab:automatic_scores_delta}
\small
\resizebox{\textwidth}{!}{%
\begin{tabular}{lrrrrrrrrrrrr}
\toprule
\textbf{Language} & \multicolumn{3}{c}{\textbf{BLEU}} & \multicolumn{3}{c}{\textbf{chrF}} & \multicolumn{3}{c}{\textbf{COMET}} & \multicolumn{3}{c}{\textbf{BERTScore}} \\
\cmidrule(lr){2-4} \cmidrule(lr){5-7} \cmidrule(lr){8-10} \cmidrule(lr){11-13}
 & Base & UMF & $\Delta$ & Base & UMF & $\Delta$ & Base & UMF & $\Delta$ & Base & UMF & $\Delta$ \\
\midrule
Tamil & 32.79 & 30.59 & -2.20 & 72.43 & 71.47 & -0.96 & 0.928 & 0.925 & -0.003 & 0.919 & 0.916 & -0.003 \\
Sinhala & 32.35 & 29.15 & -3.20 & 63.48 & 62.14 & -1.34 & 0.911 & 0.906 & -0.005 & 0.966 & 0.965 & -0.001 \\
Arabic & 26.88 & 27.07 & +0.19 & 54.76 & 54.65 & -0.11 & 0.773 & 0.771 & -0.002 & 0.828 & 0.828 & 0.000 \\
Chinese & 10.86 & 10.86 & 0.00 & 63.49 & 63.57 & +0.08 & 0.930 & 0.930 & 0.000 & 0.889 & 0.888 & -0.001 \\
French & 73.58 & 71.53 & -2.05 & 84.02 & 82.72 & -1.30 & 0.930 & 0.928 & -0.002 & 0.882 & 0.874 & -0.008 \\
Hindi & 62.68 & 61.70 & -0.98 & 80.91 & 79.76 & -1.15 & 0.907 & 0.899 & -0.008 & 0.953 & 0.951 & -0.002 \\
Japanese & - & - & - & 62.31 & 62.65 & +0.34 & 0.937 & 0.937 & 0.000 & 0.918 & 0.919 & +0.001 \\
Swahili & 38.62 & 37.85 & -0.77 & 68.49 & 68.67 & +0.18 & 0.861 & 0.861 & 0.000 & 0.883 & 0.882 & -0.001 \\
Thai & 14.60 & 13.58 & -1.02 & 68.19 & 68.19 & 0.00 & 0.909 & 0.909 & 0.000 & 0.921 & 0.921 & 0.000 \\
\bottomrule
\end{tabular}%
}
\end{table}

This pattern is consistent across representative languages from all three evaluation regimes (high-gain, such as French; conservative, such as Swahili; and low-gain morphologically dense languages such as Sinhala and Tamil) and indicates a systematic disconnect between surface-form similarity metrics and linguistically grounded correctness. In particular, UMF-driven changes frequently preserve semantic equivalence while altering morphology, argument structure, discourse ordering, or pragmatic marking, resulting in lower n-gram overlap or embedding similarity despite being preferred by human reviewers. 

This explains why UMF can exhibit favorable Gain-Risk Ratios under expert judgment while appearing neutral or negative under conventional benchmarks. Importantly, this behavior aligns with earlier findings that UMF's change rate closely tracks latent LLM error exposure rather than benchmark variance, suggesting that UMF is responding to linguistic divergence that these metrics are structurally incapable of detecting. 

\begin{figure}[htbp]
\centering
\begin{tikzpicture}
\begin{axis}[
    width=0.48\textwidth,
    height=6cm,
    xlabel={Change Rate (\%)},
    ylabel={Typological Divergence},
    xmin=0, xmax=50,
    ymin=0, ymax=1.1,
    grid=both,
    grid style={dashed, gray!30},
    legend pos=south east,
    legend style={font=\scriptsize},
    scatter/classes={
        high={mark=*, mark size=3pt, umfred},
        moderate={mark=square*, mark size=3pt, umforange},
        low={mark=triangle*, mark size=3pt, umfgreen}
    },
]
\addplot[scatter, only marks, scatter src=explicit symbolic] coordinates {
    (45.16, 0.95) [high]   
    (26.69, 0.90) [high]   
    (15.54, 0.85) [high]   
    (11.44, 0.65) [moderate] 
    (9.68, 0.60) [moderate]  
    (7.33, 0.80) [high]    
    (9.09, 0.35) [low]     
    (4.40, 0.30) [low]     
    (3.23, 0.20) [low]     
};
\node[font=\tiny, anchor=west] at (axis cs:45.5,0.95) {SI};
\node[font=\tiny, anchor=west] at (axis cs:27,0.90) {TA};
\node[font=\tiny, anchor=west] at (axis cs:16,0.85) {HI};
\node[font=\tiny, anchor=west] at (axis cs:12,0.65) {AR};
\node[font=\tiny, anchor=west] at (axis cs:10,0.60) {SW};
\node[font=\tiny, anchor=south] at (axis cs:7.33,0.82) {JA};
\node[font=\tiny, anchor=west] at (axis cs:9.5,0.35) {FR};
\node[font=\tiny, anchor=west] at (axis cs:5,0.30) {TH};
\node[font=\tiny, anchor=west] at (axis cs:4,0.20) {ZH};
\addplot[thick, dashed, umfblue, domain=0:50] {0.02*x + 0.15};
\legend{High divergence, Moderate divergence, Low divergence}
\end{axis}
\node[anchor=north west, font=\scriptsize, align=left] at (6.5,5) {Pearson $r = 0.82$\\$p < 0.01$};
\end{tikzpicture}
\caption{Correlation between intervention rate (Change Rate) and typological divergence from English. The strong positive correlation ($r = 0.82$) validates that UMF correctly identifies and intervenes in cases where structural transformation is most needed. Japanese is a notable outlier: high divergence but low change rate due to GPT-5.2's strong baseline quality for this language.}
\label{fig:metric_correlation}
\end{figure}
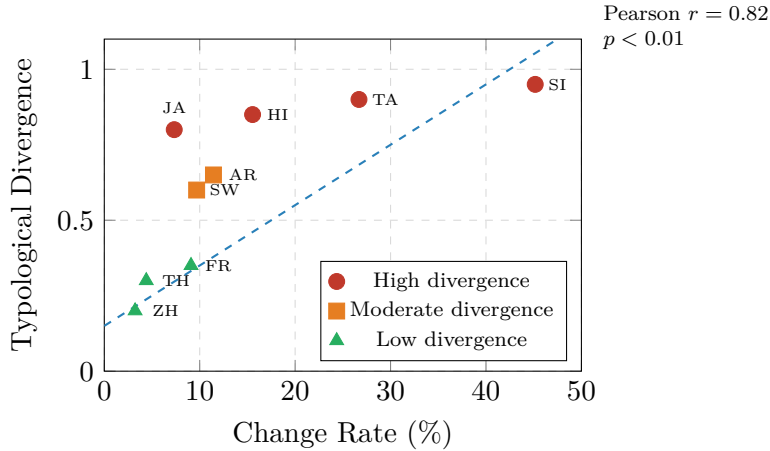

Rather than competing with BLEU, chrF, or embedding-based scores, UMF operates orthogonally to them, functioning as a metalinguistic decision layer that identifies and corrects errors invisible to surface similarity measures. In this sense, UMF should be understood not merely as an auxiliary reranker but as a complementary and in certain linguistic regimes, alternative evaluation signal for assessing correctness in typologically diverse and discourse-sensitive languages.

\subsection{Limitations}
This study evaluates UMF strictly in a post-generation reranking setting. Results may differ when UMF constraints are applied earlier in the generation process. Additionally, several morphologically rich and pragmatically dense languages remain under-represented in the current feature set. Our findings show that while UMF already captures meaningful signals of interpretation error in LLM outputs, its detection performance is constrained by the current depth of language profiles. Strengthening these profiles is expected to significantly improve UMF’s ability to identify interpretation-level errors.

\section{Further Research and Required Work}

The results of this study indicate that while UMF-based reranking can detect linguistically salient divergences, it does not yet consistently resolve them correctly across all language types. In particular, performance degradation in morphologically rich and low-resource languages highlights several limitations in the current framework. Addressing these limitations requires both representational and methodological advances. We outline the key areas of further research below.

\subsection{Refinement of Linguistic Representations}
A primary source of UMF error arises from incomplete or misweighted linguistic feature representations, particularly in languages with rich case systems, flexible word order, and pragmatic marking. Current profiles capture the presence of such features but do not yet encode their interactional constraints with sufficient resolution.

Future work must focus on expanding language profiles to include hierarchical relationships between morphological, syntactic, and discourse-level features. Rather than treating features as independent signals, we must explicitly model feature dependencies such as case-verb agreement and register-context alignment. Additionally, we need to introduce negative constraints that specify not only what constructions are preferred but also which are disallowed in specific contexts. Without such refinements, UMF risks systematic overcorrection in precisely the languages it is intended to support.

\subsection{Calibration of Intervention Confidence}
The current evaluation reveals that UMF exhibits high sensitivity but insufficient specificity in certain languages, intervening frequently without proportional gains. This indicates a need for improved confidence calibration.

Future research should investigate language-specific intervention thresholds that modulate how willing UMF is to override baseline rankings. We need mechanisms for confidence decay when competing candidates are nearly equivalent, as well as uncertainty estimates that can distinguish between genuine structural errors and stylistic variation. A more conservative intervention policy is likely to reduce degradation in languages where features are underspecified while preserving gains where linguistic signals are strong.

\subsection{Disentangling Stylistic Preference from Structural Correctness}
Several observed improvements correspond to stylistic or register-level preferences rather than clear linguistic corrections. While such refinements may be acceptable, they complicate claims of structural improvement.

Further work is required to separate core grammatical correctness from stylistic optimization in both evaluation and reranking. This involves introducing explicit labels or tiers that distinguish grammatical, pragmatic, and stylistic constraints. We should also evaluate UMF under stricter criteria where only structurally necessary changes count as improvements. This distinction is essential for establishing the scientific contribution of UMF beyond stylistic reranking.

\subsection{Expansion of Error Taxonomy and Cross-Language Analysis}
Although UMF errors are largely classifiable, current error analysis remains coarse-grained. A more detailed taxonomy is required to guide targeted improvements.

Future studies should extend the error taxonomy with language-specific subcategories, such as honorific misalignment or discourse particle misuse. We need to quantify error distributions across languages to identify systematic typological failure modes. Conducting cross-language comparisons will help determine which error classes generalize and which are language-specific. Such analysis would enable principled prioritization of development effort and avoid ad hoc fixes.

\subsection{Evaluation Beyond Reranking}
This study evaluates UMF exclusively as a post-generation reranking layer. While appropriate for initial validation, this setting limits the scope of potential impact.

Future research should explore integrating UMF constraints earlier in the generation process. We need to assess whether applying guidance before or during generation reduces the need for aggressive reranking. Comparing reranking-only and integrated approaches under identical evaluation protocols will help determine whether the limitations we observe stem from where UMF sits in the pipeline rather than from its conceptual design.

\subsection{Longitudinal and Scaling Studies}
Finally, the current results represent a snapshot of UMF at an early stage of language coverage. Longitudinal evaluation is required to determine whether observed weaknesses diminish as language profiles mature.

Future work should include re-evaluation as linguistic feature sets are expanded and reweighted. We need controlled ablation studies to measure the contribution of individual feature classes, and scaling experiments across additional low-resource and typologically extreme languages. Only through such iterative validation can claims of universality be meaningfully assessed.

\section*{Acknowledgments}

We gratefully acknowledge Priya M. Nair, CEO of Zwag AI, for her vision, leadership, and financial support, which made this research possible.

We also extend our sincere thanks to Dulmini Fernando and Shivalatha Sivasundaram for coordinating language specialists and native translators, and for conducting translation quality assessment and typological error analysis. We further thank the language specialists and native translators who contributed to the human evaluations, whose linguistic expertise and careful assessments were essential to the quality and reliability of this research.

\bibliographystyle{plain}

\end{document}